\documentclass[conference]{IEEEtran}
\IEEEoverridecommandlockouts
\usepackage{cite}
\usepackage{amsmath,amssymb,amsfonts}
\usepackage{graphicx}
\usepackage{subfigure}
\usepackage{textcomp}
\usepackage{xcolor}
\usepackage{algorithm,algorithmic}
\usepackage{bm}
\usepackage{booktabs}
\usepackage{multirow}
\usepackage{geometry}
\newgeometry{
  top=72pt,
  bottom = 54pt,
  left = 54pt,
  right = 54pt
}
\def\BibTeX{{\rm B\kern-.05em{\sc i\kern-.025em b}\kern-.08em
    T\kern-.1667em\lower.7ex\hbox{E}\kern-.125emX}}
\begin{document}
\title{Learning Task-Aware Energy Disaggregation: a Federated Approach
}

\author{\IEEEauthorblockN{Ruohong Liu}
\IEEEauthorblockA{\textit{Artificial Intelligence Thrust} \\
\textit{The Hong Kong University of Science and Technology (Guangzhou)}\\
Guangzhou, China \\
rliu519@connect.hkust-gz.edu.cn}
\and
\IEEEauthorblockN{Yize Chen}
\IEEEauthorblockA{\textit{Computational Research Division} \\
\textit{Lawrence Berkeley National Laboratory} \\
Berkeley, CA, USA \\
yizechen@lbl.gov}
}

\maketitle

\begin{abstract}
We consider the problem of learning the energy disaggregation signals for residential load data. Such a task is referred as non-intrusive load monitoring (NILM), and in order to find individual devices' power consumption profiles based on aggregated meter measurements, a machine learning model is usually trained based on large amount of training data coming from a number of residential homes. Yet collecting such residential load datasets requires both huge efforts and customers' approval on sharing metering data, while load data coming from different regions or electricity users may exhibit heterogeneous usage patterns. Both practical and privacy concerns make training a single, centralized NILM model challenging. In this paper, we propose a decentralized and task-adaptive learning scheme for NILM tasks, where nested meta-learning and federated learning steps are designed for learning task-specific models collectively. Simulation results on benchmark dataset validate proposed algorithm's performance in efficiently inferring the appliance-level consumption for a variety of homes and appliances. The code for this work is at https://github.com/RuohLiuq/FedMeta.git.
\end{abstract}

\vspace{-5pt}
\section{Introduction}
Recent electrification and proliferation of a variety of electric appliances, such as electric vehicles (EV) and IoT devices, call for advanced analytic and control tools for electric demands. With fine-grained information on individual devices' energy consumption, system operators are given the promises of designing mechanisms for achieving home energy management \cite{gopinath2020energy} and demand response~\cite{albadi2007demand} via load shifting or time-of-use pricing. 
However, most present metering systems collect subsystem-level data. For instance, typical residential-level smart meters record the total power usage including cooling and heating, safety, water, lighting, and similarly combined subsystems~\cite{mariano2021review}, and seldom include appliance-level data. To enable more accurate load analytics and more efficient demand-side management, non-intrusive load monitoring (NILM) has been proposed to improve the interactions between end users and system operators by disaggregating the total load signals into appliance-level information~\cite{hart1992nonintrusive}. 

Typical NILM techniques can be divided into two categories of tasks~\cite{alcala2017event}. The first kind of NILM algorithms tries to solve event detection problems \cite{alcala2017event, hamdi2020new}, which are designed to predict appliances' ON/OFF states over time; while other works directly predict  appliance-level power consumption \cite{kaselimi2019bayesian, kelly2015neural, d2019transfer}. Algorithms using expert heuristics \cite{meehan2014efficient}, constructing probabilistic models \cite{wild2015new, pereira2017developing}, and applying matched filters \cite{alcala2017event} have been proposed for event detection problems. However,  methods like expert heuristics depend on pre-defined threshold, which is not generalizable nor reliable considering customer behaviors and appliances' distinct physical characteristics. Probabilistic models and matched filters have similar drawbacks, as they require pre-defined template signals to match given power events. Among non-event energy disaggregation approaches, hidden Markov models (HMM) are proven to be practical in test cases \cite{bonfigli2017non, kolter2012approximate}, but HMM rely on complicated hypotheses and domain knowledge collected from data measurements \cite{d2019transfer}.

Recent advancements in deep learning have motivated researchers to adopt more complex models and larger training datasets to fulfill NILM tasks. Learning-based methods can either treat NILM as a multi-label classification problem or directly predict appliance-level signals  from aggregated inputs. \cite{kelly2015neural} applies long-short-term-memorty~(LSTM) networks for NILM tasks, as it is found that the LSTM model performs empirically better than combinatorial optimization and factorial hidden Markov models on benchmarking tests. A sequence-to-point (seq2point) model is proposed to predict single-step and appliance-level signal decomposing aggregated signal in~\cite{d2019transfer}. In \cite{2020Semisupervised} semi-supervised learning framework is proposed to cope with the challenge of inadequate labeled metering data. We refer the readers to \cite{hosseini2017non} for a review of NILM techniques.

\begin{figure}[htbp]
\centerline{\includegraphics[width=0.45\textwidth]{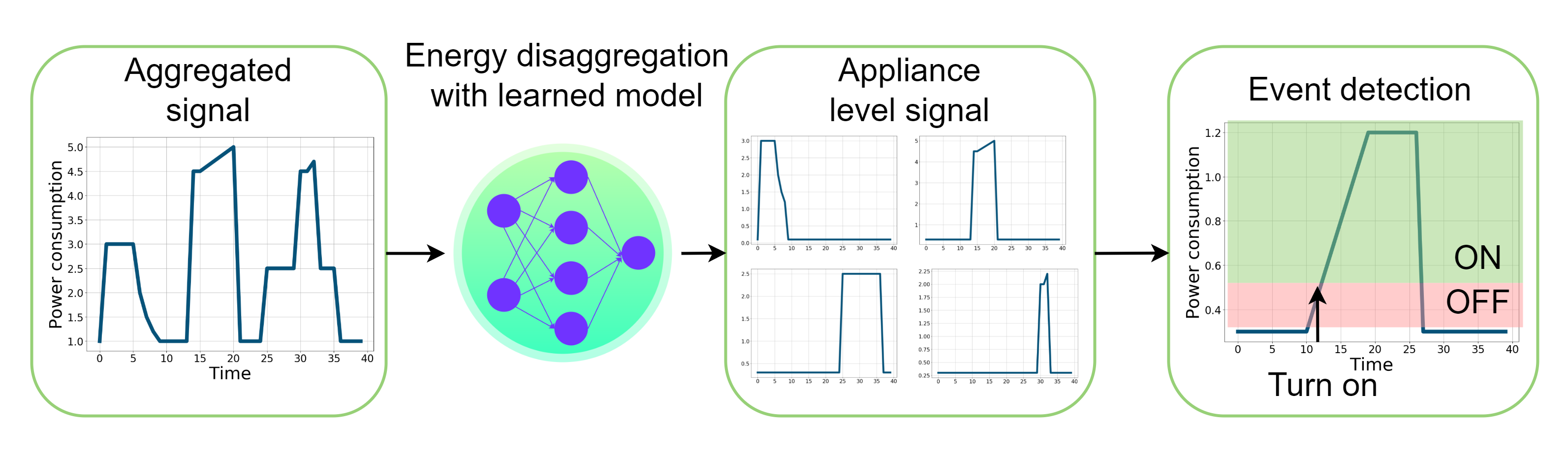}}
\vspace{-7pt}
\caption{Overview of NILM task. Once trained, machine-learning based model is able to output appliance-level power consumption profiles, which can be further processed to determine the ON/OFF Status.}
\label{fig1}
\end{figure}
\vspace{-7pt}

\newgeometry{
  top=54pt,
  bottom = 54pt,
  left = 54pt,
  right = 54pt
}
However, there is growing recognition that the success of current machine learning techniques relies on large and comprehensive data, while it is challenging for trained models to adapt to heterogeneous data distributions with a variety of energy consumption behaviors and regional characteristics. Meanwhile, most of the proposed algorithms train NILM models in a centralized manner, requiring data of all users to be gathered together. Such practice may also raise concerns over smart meter data privacy~\cite{lin2021privacy}. Moreover, such data-driven approaches have high demands on computing resources due to data communication and training needs, which is challenging for the current power grid infrastructure. 

In this work, we are interested in utilizing the representation capability of deep learning models while addressing the challenges of data collection/communication and load data heterogeneity using one algorithm. We achieve this by collaboratively training a group of meta-learned models in a federated approach. Federated learning involves training statistical models using local devices with local data \cite{li2020federated}. With federated learning, residential communities can train a neural network locally using their metering data and only upload the learned model parameters to the server for aggregation. Thus, the server can learn a global model by aggregating the local model parameters without knowing local data. In addition, meta learning can help learned models quickly adapt to new distribution of NILM datasets. In energy-related applications, meta-learning makes it possible to learn characteristics of building models from multi-source data~\cite{zhanmeta}. Our work is motivated by \cite{jiang2019improving, chen2018federated}, and we utilize the close relationship between federated learning update steps and model-agnostic meta learning update steps~\cite{finn2017model}. We design a unified framework tailored for NILM tasks, \texttt{FedMeta}, to achieve both fast decentralized algorithm training and accurate task-adaptive localized energy disaggregation. We can not only disaggregate smart meter signals using \texttt{FedMeta}, but also achieve event detection for a number of appliances as presented in Fig.~\ref{fig1}. As far as the authors are aware, the proposed method is the first attempt to achieve a balance of metering data privacy and disaggregation performance, with simulation results validating algorithm performance on real-world residential load data under various settings.

\vspace{-5pt}
\section{Problem Formulation and Preliminaries}
In this section, we start from describing the formulation of NILM task. We will then discuss the challenges coming from data collection side and model adaptation, which motivates us to design the \texttt{FedMeta} algorithm.

\vspace{-3pt}
\subsection{NILM Problem Setup}\label{AA}
For electric utilities, system operators implement NILM to extract appliance-level power consumption signals from disaggregating meter signals. We assume that the window size is $2T$ and we have $A$ individual possible appliances. Let $y_{t}$ denote the raw meter readings in a household at time t, and $x_{t}^{j}$ represents the power reading of the $j$-th appliance in the household. Then $y_{t}$ can be represented as:
\begin{equation}
    y_{t} = \sum_{j=1}^{A}x_{t}^{j}+e_{t},
\end{equation}
where $e_t$ denotes noisy signals.

In our learning-based NILM model, we apply a sequential model to map a fixed-length time series of meter readings to the midpoint of a group of individual appliances \cite{d2019transfer}. The model input at time step $t$ can be written as $\mathbf{y}_{t:t+2T-1} = [y_{t},y_{t+1},...,y_{t+2T-1}]$, while the output of model is $\mathbf{x}_{t+T} = [x_{t+T}^{(1)},x_{t+T}^{(2)},...,x_{t+T}^{(A)}]$. We are interested in finding a function $f$ to get $\mathbf{x}_{t+T}$ based on $\mathbf{y}_{t:t+2T-1}$:
\begin{equation}
    f(\mathbf{y}_{t:t+2T-1}) = \mathbf{x}_{t+T}.
\end{equation}
Without loss of generality, we may omit the timestep subscript of $\mathbf{x}$ and $\mathbf{y}$ hereinafter. Since we are able to collect historical samples $\{\mathbf{y}, \mathbf{x}\}$, it is possible to fit $f$ in a supervised approach. Let the vector $\mathbf{w} \in \mathbb{R}^{d}$ denote the model parameters of $f$, and let $\mathcal{L}(\mathbf{w})$ denote the mean-squared-error (MSE) loss given $\mathbf{w}$:
\begin{equation}
    \mathcal{L}(\mathbf{w}) = \mathbb{E}(f(\mathbf{y})-\mathbf{x})^{2}.
\end{equation}

With the decomposed function and decomposed values, we can also detect appliances' ON/OFF state over time. It is possible to choose a  predefined threshold for each appliance. When the power consumption of the appliance in a certain time step is greater than the threshold, we infer that appliance is ON. Previous studies focus on designing better machine learning models to represent the complex temporal patterns and contextual information in the aggregated signals. Typically, it is assumed that central model with data collected from a number of customers is trained. In the following subsections, we will describe the evolving challenges regarding both training paradigms and task differences.

\subsection{Federated Learning Model}
\label{sec:FL}
To achieve desired performance in NILM task, data-driven approaches require large datasets to train the models. However, both the power consumers and system operators in a competitive electricity market may be unwilling to share their raw data due to privacy concerns \cite{lin2021privacy}. Meanwhile, local devices may suffer from constraints in storage, computation, and communication capacities \cite{chen2018federated}. Such challenges can be potentially tackled with federated learning~(FL) based on decentralized datasets and training schemes. \texttt{FedAvg} is an algorithm based on iterative model averaging that has been demonstrated to be effective in computer vision tasks \cite{mcmahan2017communication}. In this paper, we refer such setting as \texttt{FedAvg}, and take it as a benchmark algorithm to compare with our \texttt{FedMeta} algorithm introduced in Section \ref{simulation section}.

In federated learning, local devices are referred to as clients. Each client has its own dataset to train a local model, and then shares model parameters rather than data with central server, which is in charge of training a global model. In our federated learning setting, suppose there are $M$ clients representing different cluster of customers, and each of them possesses historical data samples $D_{i}=\{\textbf{y}_i, \textbf{x}_i\}$. Thus, the objective function of each client can be written as $\mathcal{L}^{(i)}(\mathbf{w})= \mathbb{E}_{(\textbf{y},\textbf{x})\sim D_{i}}[\mathcal{L}(\mathbf{w})]$. Consider a set of objective functions: $\{\mathcal{L}^{(i)}: \mathbf{w}^{(i)} \in \mathbb{R}^d\}_{i=1}^{M}$ for the $i=1,...,M$. For federated learning, we are then interested in learning a model $\mathbf{w}^{FL}$ collectively by all clients:
\begin{equation}
\label{equ:fl}
    \min _{\mathbf{w}^{FL} \in \mathbb{R}^{d}}\left[\mathcal{L}(\mathbf{w}):=\sum_{i=1}^{M} \alpha^{(i)} \mathcal{L}^{(i)}(\mathbf{w}^{(i)})\right],
\end{equation}
where $\alpha^{(i)}$ is a weighting parameter for each client participating in the NILM task. Normally, we choose $\alpha^{(i)}$ based on the number of data samples we have for each client.

To iteratively solve the training problem \eqref{equ:fl}, the gradient step of federated learning is as follows. Firstly, at iteration $k$, clients download the global model from the server, and implement a certain number of local update steps with gradient step $\gamma$: 
\begin{equation}
\label{equ:FL_update}
    \mathbf{w}_{k+1}^{(i)}=\mathbf{w}_{k}^{(i)}- \gamma\nabla\mathcal{L}^{(i)}(\mathbf{w}_{k}^{(i)}).
\end{equation}
Next, all clients send their model parameters $\mathbf{w}^{(i)}$ to the server, and the server then calculates a weighted average of local model parameters:

\vspace{-10pt}
\begin{equation}
    \mathbf{w}_{k+1}=\sum_{i=1}^{M} \alpha^{(i)}\mathbf{w}_{k+1}^{(i)}.
\end{equation}
\vspace{-10pt}

With such a iterative update procedure, \texttt{FedAvg} achieves the decentralized training goal without explicit data sharing from each power consumer. Yet such a model still does not consider the inherent differences between customers' NILM tasks, and the fitted disaggregating function $f(\mathbf{y})$ may not be the optimal fit for all NILM clients.

\subsection{Meta-Learning for Task Adaptation}\label{2C}
Federated learning provides an ideal candidate to make full use of datasets from all residential homes while achieving decentralized model updates. However, to achieve desired results in various homes and communities, we need to consider the differences of models between regions. Thus we want to find a way to adjust the model locally. To achieve this, Model-agnostic meta-learning (MAML) is a possible fit, as MAML algorithm effectively bootstraps from a set of tasks to learn faster on a new task~\cite{finn2017model}. Distinct to transfer learning, the goal of meta-learning is to train a group of learnable models collectively instead of training a pre-trained model.

In MAML, the notion of task is closely related to the client in federated learning, as each task owns a local dataset. The goal for each task is to train a local model given this specific dataset, while the dataset in a task can be much smaller than that in a client. Regarding our NILM problem, a task always contains power data of a household in a particular month, while a client may have data for a house or even all houses in a region for up to one year. We assume these task data are drawn from a fixed distribution, $ \mathcal{T}_{i}\sim\mathbb{P}(\mathcal{T})$. 

In meta training step, MAML attempts to learn an initial set of parameters $\mathbf{w}$ using $N$ tasks $\{\mathcal{T}_{i}\}_{i=1}^{N}$, and each task includes a labeled dataset $\mathcal{D}_{i}$. At iteration $f$, we adapt the model $f(\mathbf{y})$ with model parameters $\mathbf{w}_{f}$ to each task by updating the model parameters $\mathbf{w}_{f}$ to $\mathbf{w}_{f+1}^{(i)}$. In this work, a 1-step gradient update procedure is applied to calculate $\mathbf{w}_{f+1}^{(i)}$ as follows:
\begin{equation}
    \mathbf{w}_{f+1}^{(i)}=\mathbf{w}_{f}-\gamma\nabla\mathcal{L}^{(i)}(\mathbf{w}_{f}),
\end{equation}
where $\gamma$ denotes the learning rate.

Remember that with meta-learning, we want the model to perform well across all sampled tasks after a 1-step gradient update from $\mathbf{w}_{f}$. In other words, we aim to minimize the summation of MSE loss of all tasks computed by updated parameters $\mathbf{w}_{f+1}^{(i)}$:

\vspace{-5pt}
\begin{equation}
\label{equ:maml}
    \min _{\mathbf{w}_{f}}\frac{1}{N}\sum_{i=1}^{N}  \mathcal{L}^{(i)}(\mathbf{w}^{(i)}_{f+1}),
\end{equation}
\vspace{-5pt}

In order to solve the optimization problem in (8), we update $\mathbf{w}$ with $\mathcal{L}^{(i)}$ as follows:

\vspace{-5pt}
\begin{equation}
\label{equ:MAML_update}
    \mathbf{w}_{f+1} = \mathbf{w}_{f} - \beta \nabla\sum_{i=1}^{N}  \mathcal{L}^{(i)}(\mathbf{w}_{f+1}^{(i)}),
\end{equation}
\vspace{-5pt}

where we use $\beta$ to denote the stepsize of MAML. Note that in (9), the optimization variable is initial parameter $\mathbf{w}_{f}$, while MSE loss is collected from all training tasks with updated parameters $\mathbf{w}_{f+1}^{(i)}$.

\vspace{-5pt}
\begin{figure}[htb]
\centerline{\includegraphics[width=0.4\textwidth]{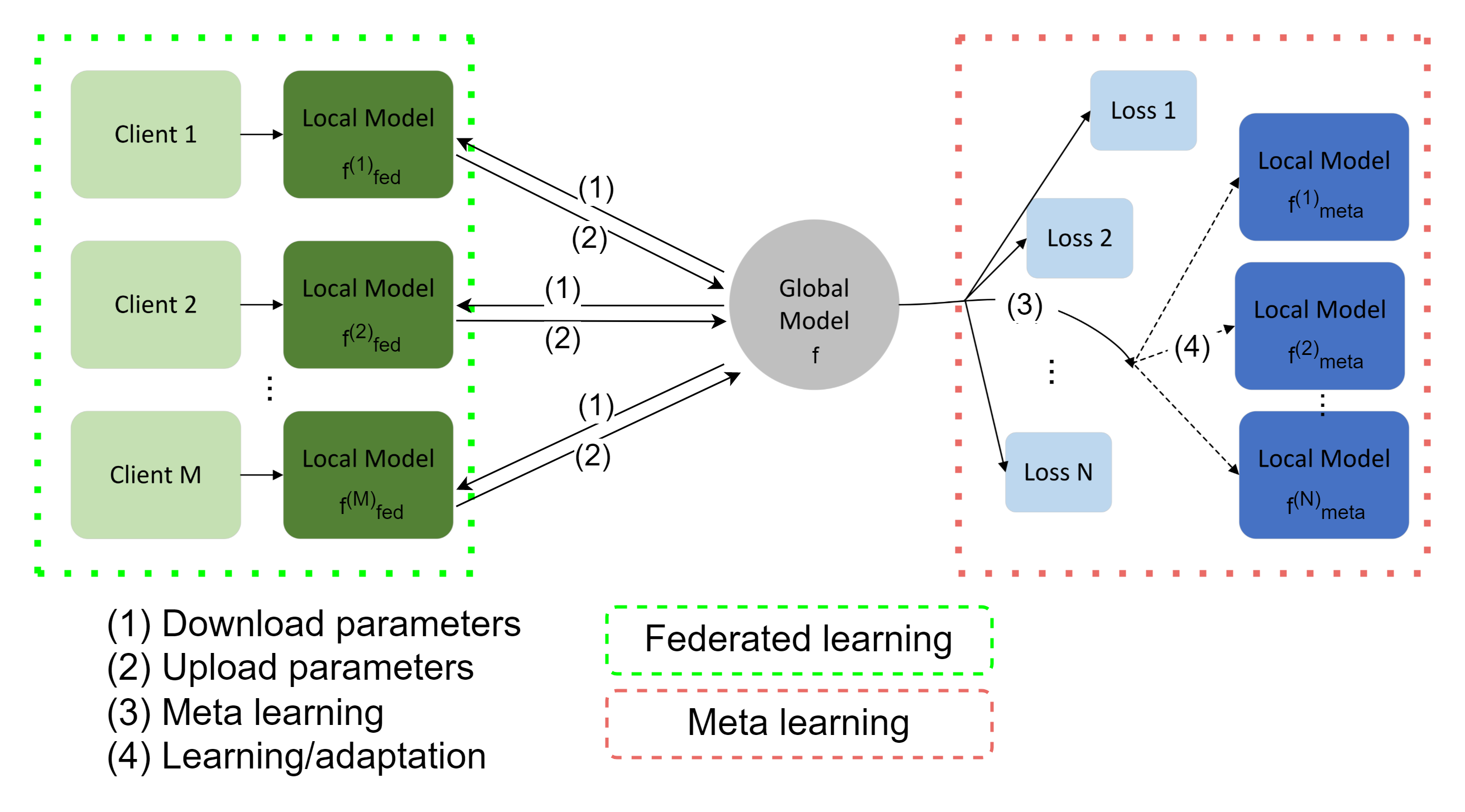}}
\vspace{-5pt}
\caption{The workflow of proposed \texttt{FedMeta} algorithm for NILM task. We combine both federated learning and meta-learning to optimize an initial model (global model) that can quickly adapt to new testing tasks.}
\label{fig2}
\end{figure}

Next, in meta testing step, we will test the ability of initial parameters to adapt to new tasks. Therefore, we need to do a few fine-tune steps starting from the initial parameters we trained. 


The intuition behind designing a meta learner for NILM task is each cluster of electricity users has distinct load behaviors, and MAML model is able to extract and propagate representations of aggregated load data across different communities and homes. This would later boost the \texttt{FedMeta} algorithm performance for task-specific load disaggregation.

\vspace{-5pt}
\section{Fed-Meta NILM Algorithm}
In this section, we describe the design principles of \texttt{FedMeta} algorithm and illustrate how the algorithm works on NILM tasks. The detailed model architecture for predicting decomposed signal for NILM will be illustrated in Section \ref{inplementation details}.

\begin{algorithm}[htb]
\label{algorithm1}
\caption{FedMeta Algorithm for NILM}
\begin{algorithmic}[1]
\REQUIRE 
$T$:main loop iterations; $K$:Federated learning iterations; 
$F$:MAML iterations; $\mathbf{w}$:initial model parameters;
$M$:number of clients;
$N$:number of tasks;
$\beta$:MAML learning rate;\\
\STATE // train initial model
\STATE Randomly initialize $\mathbf{w}_{global}$
\FOR{each round $t=1,2,...,T$}
\STATE $\mathbf{w}_{1} \gets \mathbf{w}_{global}$
\FOR{each FL round $k=1,2,...,K$}
\STATE $ S_{k} \gets$ (random set of M clients)
\FOR{each client $i \in S_{k}$}
\STATE $\mathbf{w}_{k+1}^{(i)}\gets \mathbf{LocalUpdate}(\mathbf{w} _{k})$
\ENDFOR
\STATE $\mathbf{w}_{k+1} \gets\sum_{i=1}^{M}\frac{1}{M}\mathbf{w}_{k+1}^{(i)}$
\ENDFOR
\STATE $\mathbf{w}_{global} \gets \mathbf{w}_{K}, \mathbf{w}_{1} \gets \mathbf{w}_{global}$
\FOR{each MAML round $f=1,2,...,F$}
\STATE $S_{f} \gets$ (random set of N tasks)
\FOR{each task $i \in S_{f}$}
\STATE $\mathbf{w}_{f+1}^{(i)}\gets \mathbf{LocalUpdate}(\mathbf{w} _{f})$
\ENDFOR
\STATE Update $\mathbf{w}_{f+1} \gets \mathbf{w}_{f} - \beta\nabla\sum_{i=1}^{N}\mathcal{L}^{(i)}(\mathbf{w} _{f+1}^{(i)})$
\ENDFOR
\STATE $\mathbf{w}_{global} \gets \mathbf{w}_{F}$
\ENDFOR
\STATE // fine-tune for testing task
\STATE $\mathbf{w}_{test}\gets \mathbf{LocalUpdate}(\mathbf{w}_{global})$

\REQUIRE 
$\gamma$: client/task learning rate; $E$: client update epoches;\\
\STATE $\mathbf{function}\enspace \mathbf{LocalUpdate} (\mathbf{w})$
\STATE $\enspace\enspace\enspace$ Split local dataset into batches B
\STATE $\enspace\enspace\enspace\enspace\enspace\enspace \mathbf{for}$ each epoch $e = 1,2,...,E\enspace\mathbf{do}$
\STATE $\enspace\enspace\enspace\enspace\enspace\enspace\enspace\enspace\enspace \mathbf{for}$ batch $b \in B\enspace\mathbf{do}$
\STATE $\enspace\enspace\enspace\enspace\enspace\enspace\enspace\enspace\enspace\enspace\enspace\enspace\mathbf{w}=\mathbf{w}- \gamma\nabla\mathcal{L}(\mathbf{w},b)$
\STATE $\enspace\enspace\enspace\enspace\enspace\enspace\enspace\enspace\enspace \mathbf{end}$ $\enspace  \mathbf{for}$
\STATE $\enspace\enspace\enspace\enspace\enspace\enspace \mathbf{end}$ $\enspace  \mathbf{for}$

\STATE $\enspace\enspace\enspace \mathbf{return}\enspace\mathbf{w}$
\label{code:recentEnd}
\end{algorithmic}
\end{algorithm}

\label{BB}
Past attempts to design federated learning or transfer learning schemes to solve the NILM problem only focus on either data sharing and training schemes or model adaptivity \cite{d2019transfer, zhang2021fednilm}. In this work, we find it is possible to address both by coordinating federated learning steps and MAML updates. Based on the initial model trained by federated learning steps, each power consumption data owner can rapidly train a localized model which is suitable for its own data with few steps of MAML updates. Driven by such motivation, we propose the \texttt{FedMeta} algorithm. 

The framework of proposed \texttt{FedMeta} learning scheme is presented in Fig. \ref{fig2}. Similar to the update procedure in Section \ref{sec:FL}, in federated learning steps, we initialize a global model and let the clients download model parameters from the global model. Each client trains its local model with parameters $\mathbf{w}^{(i)}$ with its own dataset separately, then sends model parameters back to the system operator, who is taking the role of the central agent. The global model updates $\mathbf{w}_{global}$ by calculating the weighted average of local model parameters. 
 
 Then we consider the meta training step based on Section \ref{2C}, in which we train a local model for each task to reflect each task's distinct characteristics of power consumption. Each task keeps its distinct model while contributing to the overall model update based on Equation \eqref{equ:maml}. With the total loss $\sum_{i=1}^{N}\mathcal{L}^{(i)}$ calculated, we can update the global model $\mathbf{w}_{global}$ along with task-specific model with parameters $\mathbf{w}^{(i)}$.
 
We explicitly set up the connection between federated learning steps and MAML steps given a set of clients with local data. Specifically, the update rule \eqref{equ:FL_update} for federated learning and \eqref{equ:MAML_update} for MAML can be integrated together with the same model architecture. And we implement these two learning algorithms iteratively to achieve fast, decentralized model updates and model adaptations simultaneously. Note that in both federated learning step and meta-learning step, we locally update the model with a few times of gradient updates, and both the number of updating steps and stepsize are tunable parameters. We also remark that the proposed \texttt{FedMeta} provides a more efficient and consistent training paradigm than directly applied MAML, as the learner can all contribute to the global model updates. Finally, in the meta testing stage where load data from unknown community comes, we are able to find a localized model for each task with its own dataset. By doing a few steps of local updates in each testing task, our model can easily achieve good performance in predicting appliance-level power consumption. The whole algorithm is given in Algorithm 1.


\vspace{-5pt}
\section{Numerical Studies}
\label{simulation section}
\subsection{Dataset and Task Description}

Throughout simulations, we use public residential load data from Pecan Street. They provided access to static time-series datasets for 25 homes from Austin. All of the values in these datasets are average real power over the interval. We choose the 1-minute energy datasets of 6 homes collected from Austin, which contains all of the electricity data collected by Pecan Street’s eGauge devices in 2018 \cite{parson2015dataport}. We focus on 4 appliances: electric vehicle, air compressor, dryer, and oven. The reason we chose these 4 appliances is that they show heterogeneous patterns across various homes and different seasons while making a significant portion of overall electricity usage across all timesteps.

\vspace{-10pt}
\begin{figure}[htbp]
\centerline{\includegraphics[width=0.45\textwidth]{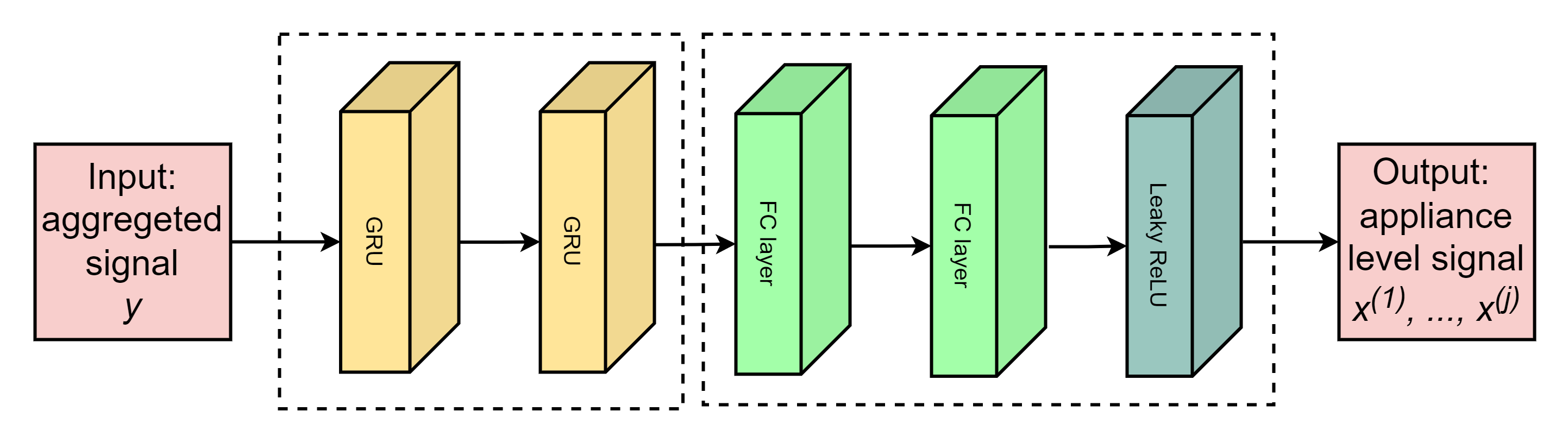}}
\caption{Neural network architecture for NILM tasks.}
\label{fignn}
\end{figure}
\vspace{-10pt}
 
We combine data from 2 homes to form the dataset of federated learning step and divide them evenly based on the number of clients. In our experiment, we have 10 clients, each of which has 12,000 training samples. We also utilize the data from 3 homes as the datasets of various tasks in meta-learning step, each of them include 12,000 training samples. As for testing step, we chose data from 3 homes to form different tasks, and each task has 12,000 training samples for fine-tune step and 24,000 samples for testing step.  We normalize the original aggregated signals to the range of $[0,1]$.



\subsection{Implementation Details}
\label{inplementation details}

\textbf{Model Architecture}: In our experiment, we use Gated Recurrent Unit (GRU) as the major neural network block, which is a kind of recurrent neural networks(RNN) with adaptive gating mechanism~\cite{chung2014empirical}. RNN can map from an input vector to an output vector while considering temporal relations, making it suitable for sequential data. GRU helps overcome the problem of 'vanishing gradient' problem of standard RNN models. We use LeakyReLU as activation function, and the maximum dimension of the neural network is 480, with an input sequence length of 120 (2 hours). The network structure is detailed in Fig. \ref{fignn}.

\vspace{-10pt}
\begin{figure}[htbp]
\centerline{\includegraphics[width=0.35\textwidth]{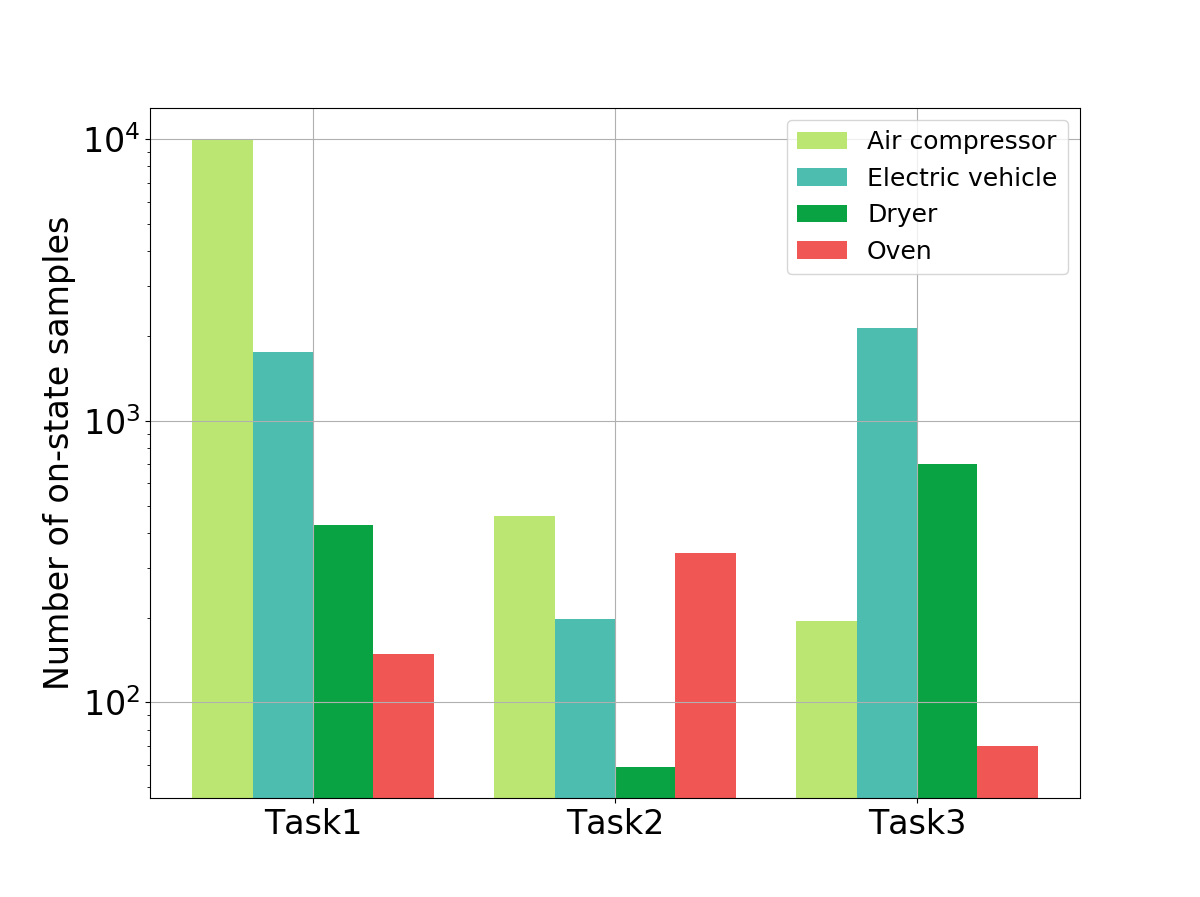}}
\caption{Data distribution of ON-state samples of 3 testing tasks. We randomly choose 3 testing tasks from NILM task pool. Each of them has equal amount of samples, while the numbers of ON-state sample are highly distinct.}
\label{distribution}
\end{figure}
\vspace{10pt}

\begin{figure*}[htbp]
\centerline{\includegraphics[width=0.85\textwidth]{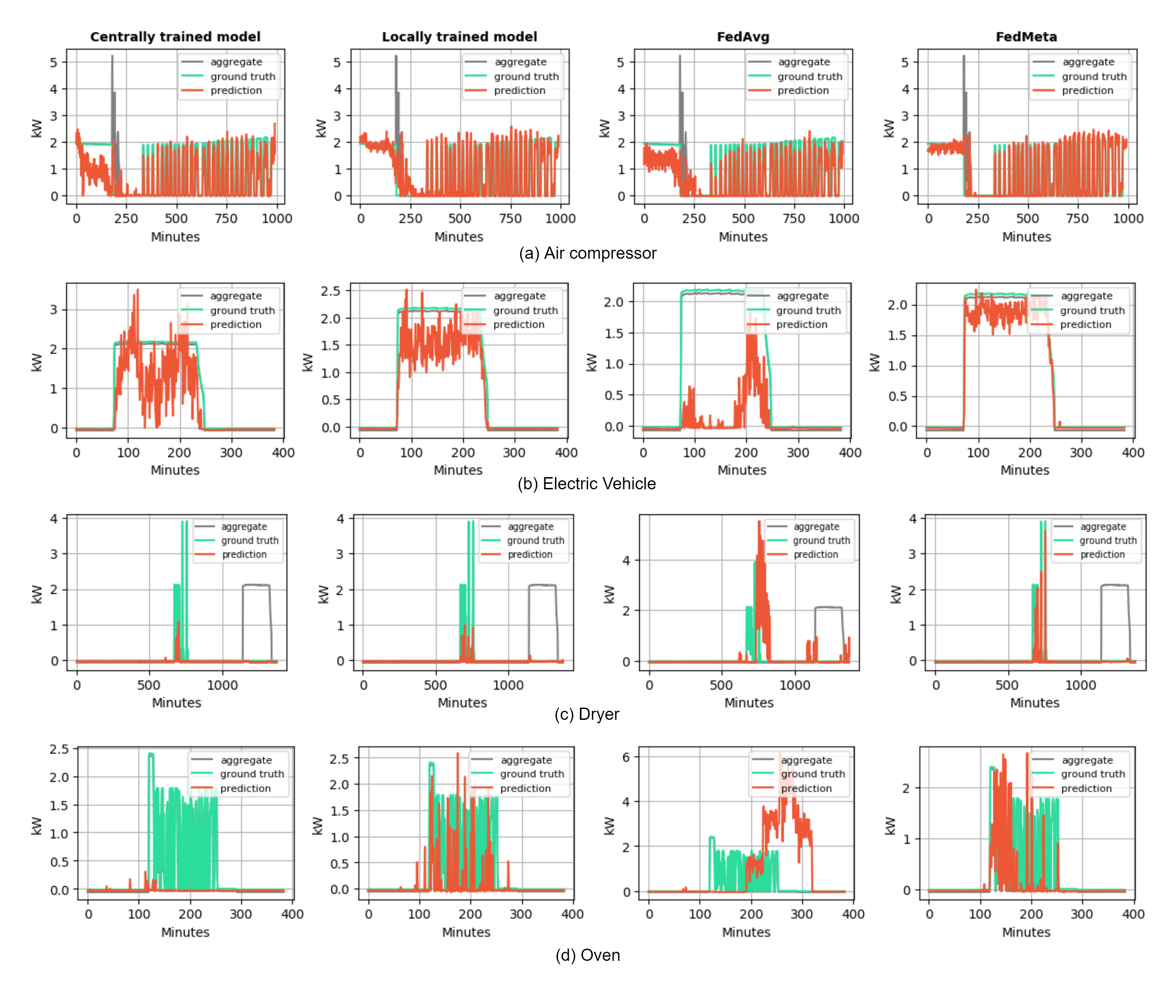}}
\caption{Energy disaggregation results of 4 appliances on testing tasks using 3 benchmark methods and proposed \texttt{FedMeta}. For each row, we show the aggregated signal of 4 appliances (grey), ground truth signal of an appliance (light blue), and disaggregated signal predictions  (red).} 
\label{plot3}
\end{figure*}
\vspace{-5pt}

\vspace{-5pt}
\begin{figure}[htbp]
\centerline{\includegraphics[width=0.45\textwidth]{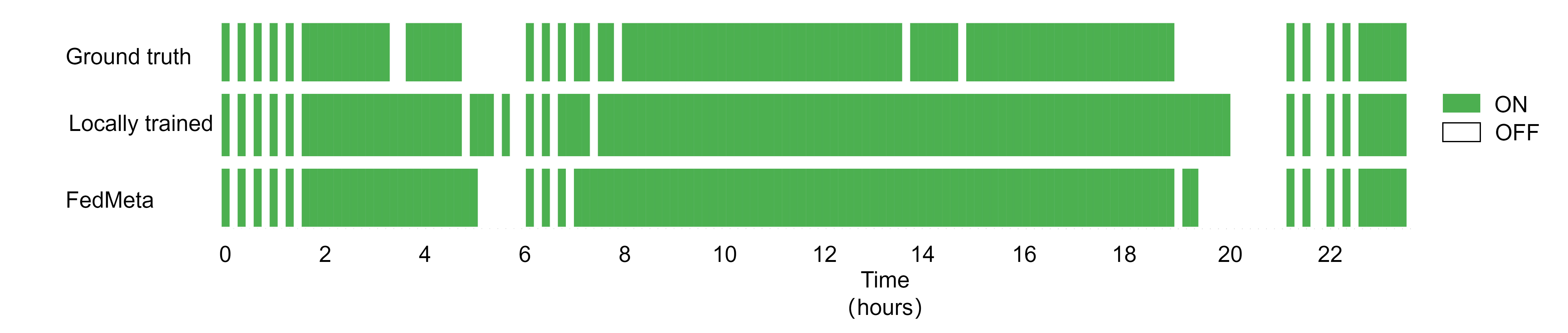}}
\caption{ON/OFF state detection of air compressor in 24 hours' testing samples with 10 minutes' intervals. Three rows show appliance state ground truth detection results of locally trained model and \texttt{FedMeta} respectively. At the 5th, 19th, and 21st hours, the air compressor is turning off. The power value decomposed by \texttt{FedMeta} are in good agreement with the ground truth.}
\label{heatmap_v2}
\end{figure}
\vspace{-10pt}

\textbf{Benchmarking Models}: To validate proposed algorithm's performance, we set the neural network unchanged, but compare the disaggregation results following different training algorithms:
\begin{itemize}
\item Centrally-trained model: This training paradigm utilizes data from all testing tasks to train a central model. Given $N$ testing tasks, we only train one model and report test results on each task;
\item Locally-trained model: For each task, the model only uses its own dataset to train a local model. Given $N$ testing tasks, we need to train $N$ local models and test them on corresponding tasks;
\item \texttt{FedAvg}: Each client trains a local model using its own dataset. Next, clients send model parameters to the server. The server then updates the global model and send new global model parameters to clients for next local update iteration. Given $M$ clients, we need to train a global model with $M$ local-updating training steps, and finally save the global model to test performance on each task.
\end{itemize}

For algorithm evaluation we include the following metrics: appliance detection accuracy, F1 score, mean absolute error (MAE) and signal aggregate error (SAE). For each appliance $j$, MAE is defined as follows:
\begin{equation}
        MAE = \frac{1}{L}\sum_{l=1}^{L}\lvert \hat{x}_{l}^{(j)}-x_{l}^{(j)} \rvert,
\end{equation}
where $L$ denotes the total number of testing samples in one task.
 SAE is defined as follows:
\begin{equation}
    SAE = \frac{\lvert \hat{E}^{(j)}-E^{(j)} \rvert}{max(E^{(j)},\hat{E}^{(j)})},
\end{equation}
where $\hat{E}= \sum_{l=1}^{L}{\hat{x}}^{(j)}_{l}$ and $E = \sum_{l=1}^{L} x_{l}^{(j)}$ refers to total predicted energy and total actual energy respectively.


\subsection{Simulation Results}

We show model performances on 3 testing tasks including different homes, and each of the task includes 4 appliances' data. Fig.~\ref{distribution} shows the distribution of working statuses for all appliances. We can see the data samples are unbalanced across all tasks, meaning that NILM models can be different for them. We compare the performance of centrally trained model, locally trained model, and models trained with \texttt{FedAvg} and  \texttt{FedMeta} on 4 appliances.



\begin{table}
\centering
\caption{Average Model performances comparisons on all tasks}
\begin{tabular}{llllll} 
\hline
Device                                                                                        & Algorithm         & F1             & Acc.           & MAE            & SAE             \\ 
\hline
\multirow{4}{*}{Air Compressor}                                                               & Centrally-trained & 0.637          & 0.935          & 0.181          & 0.356           \\
                                                                                              & Locally-trained   & 0.547          & 0.95           & \textbf{0.123} & 0.590           \\
                                                                                              & \texttt{FedAvg}   & 0.602          & 0.925          & 0.213          & 0.495           \\
                                                                                              & \texttt{FedMeta}  & \textbf{0.770} & \textbf{0.966} & 0.136          & \textbf{0.216}  \\ 
\hline
\multirow{4}{*}{Electric Vehicle}                                                             & Centrally-trained & \textbf{0.656} & 0.950          & 0.130          & 0.91            \\
                                                                                              & Locally-trained   & 0.646          & \textbf{0.964} & 0.110          & 0.675           \\
                                                                                              & \texttt{FedAvg}   & 0.396          & 0.927          & 0.163          & 0.970           \\
                                                                                              & \texttt{FedMeta}  & 0.607          & \textbf{0.958} & \textbf{0.105} & \textbf{0.672}  \\ 
\hline
\multirow{4}{*}{Dryer}                                                                        & Centrally-trained & 0.165          & 0.971          & \textbf{0.063} & 1.350           \\
                                                                                              & Locally-trained   & 0.306          & 0.971          & 0.071          & 1.167           \\
                                                                                              & \texttt{FedAvg}   & 0.110          & 0.898          & 0.172          & \textbf{0.747}  \\
                                                                                              & \texttt{FedMeta}  & \textbf{0.412} & \textbf{0.976} & 0.072          & 0.913           \\ 
\hline
\multirow{4}{*}{Oven}                                                                         & Centrally-trained & 0.022          & \textbf{0.982} & 0.038          & 1.189           \\
                                                                                              & Locally-trained   & 0.129          & 0.969          & 0.053          & 1.279           \\
                                                                                              & \texttt{FedAvg}   & 0.056          & 0.785          & 0.230          & \textbf{0.934}  \\
                                                                                              & \texttt{FedMeta}  & \textbf{0.271} & \textbf{0.980} & \textbf{0.034} & 1.146           \\
\hline
\end{tabular}
\label{table1}
\end{table}

\begin{table}[t]
\caption{Average Model performances comparisons on Task 3}
\begin{center}
\begin{tabular}{llllll} 
\hline
Device                                                                                        & Algorithm         & F1             & Acc.           & MAE            & SAE            \\ 
\hline
\multirow{4}{*}{Air Compressor}                                                               & Centrally-trained & 0.315          & 0.894          & 0.132          & 0.777          \\ 
                                                                                              & Locally-trained   & 0.032          & 0.951          & \textbf{0.078} & 1.427          \\ 
                                                                                              & \texttt{FedAvg}   & 0.254          & 0.873          & 0.193          & 0.833          \\ 
                                                                                              & \texttt{FedMeta}  & \textbf{0.660} & \textbf{0.977} & 0.161          & \textbf{0.523}  \\ 
\hline
\multirow{4}{*}{Electric Vehicle}                                                             & Centrally-trained & 0.817          & \textbf{0.962} & 0.138          & 0.478          \\ 
                                                                                              & Locally-trained   & 0.825          & 0.956          & 0.110          & 0.148          \\ 
                                                                                              & \texttt{FedAvg}   & 0.497          & 0.912          & 0.218          & 0.801    
                                                                                            \\ 
                                                                                              & \texttt{FedMeta}  & \textbf{0.835} & \textbf{0.959} & \textbf{0.102} & \textbf{0.049}   \\ 
\hline
\multirow{4}{*}{Dryer}                                                                        & Centrally-trained & 0.347          & 0.969          & 0.081          & 1.466          \\ 
                                                                                              & Locally-trained   & \textbf{0.486} & 0.968          & 0.075          & 1.050          \\ 
                                                                                              & \texttt{FedAvg}   & 0.180          & 0.902          & 0.161          & \textbf{0.314} \\ 
                                                                                              & \texttt{FedMeta}  & 0.454          & \textbf{0.970} & \textbf{0.077} & 1.259          \\ 
\hline
\multirow{4}{*}{Oven}                                                                         & Centrally-trained & 0.024          & \textbf{0.982} & \textbf{0.024} & 0.619
                                                                                            \\  
                                                                                              & Locally-trained   & 0.028          & 0.959          & 0.043          & \textbf{0.207}   \\ 
                                                                                              & \texttt{FedAvg}   & 0.010          & 0.870          & 0.207          & 1.062     
                                                                                            \\ 
                                                                                              & \texttt{FedMeta}  & \textbf{0.146} & \textbf{0.984} & \textbf{0.025} & 0.633          \\ 
\hline
\end{tabular}
\end{center}
\label{table2}
\end{table}

In Fig.~\ref{distribution}, we can observe that the sample distribution is highly unbalanced for all tasks. And the lack of positive samples may impact its disaggregation performance. On one hand, compared to locally-trained model and centrally-trained model, model trained with \texttt{FedMeta} is seldom influenced by unbalanced data of each task, for the reason that it can learn general patterns for all appliances through federated learning steps.  On the other hand, due to the task differences, a centrally-trained model and model learned with \texttt{FedAvg} may fail to achieve fair performance across all tasks. \texttt{FedMeta} overcomes such difficulty by fine-tuning a specific model for every new task starting from an initial model.

Table~\ref{table1} reports average performance of 4 algorithms on all testing tasks. 
The results presented in Table~\ref{table1} suggest that the \texttt{FedMeta} outperforms other 3 benchmark models under most metrics, and \texttt{FedMeta} shows more consistent performance for all 4 appliances. Fig.~\ref{plot3} presents the energy disaggregation time-series of 4 appliances, from which we have the following observations: (1) In all of 4 presented appliances, locally-trained model and \texttt{FedMeta} can correctly detect ON/OFF states  most of the time. The centrally-trained model fails to detect ON-state of oven, while the \texttt{FedAvg} model wrongly detects an ON-state when dryer is off; (2) Model trained with \texttt{FedMeta} outperforms other 3 benchmark models on energy disaggregation, as it can disaggregate the power signal, and consistently predicts more precise consumption value against ground truth for all appliances. Fig.~\ref{heatmap_v2} makes a comparison between locally-trained model and \texttt{FedMeta}'s performance on event detection. As is shown for this model testing data sample, locally-trained model wrongly detects two periods. In Table \ref{table2}, we illustrate the model's specific performance on task 3, 
indicating that \texttt{FedMeta} is achieving better performance in the single task compared to \texttt{FedAvg} and centrally-trained model, while the performance is comparable to locally-trained model.

\vspace{-5pt}
\begin{figure}[htb]
\centerline{\includegraphics[width=0.35\textwidth]{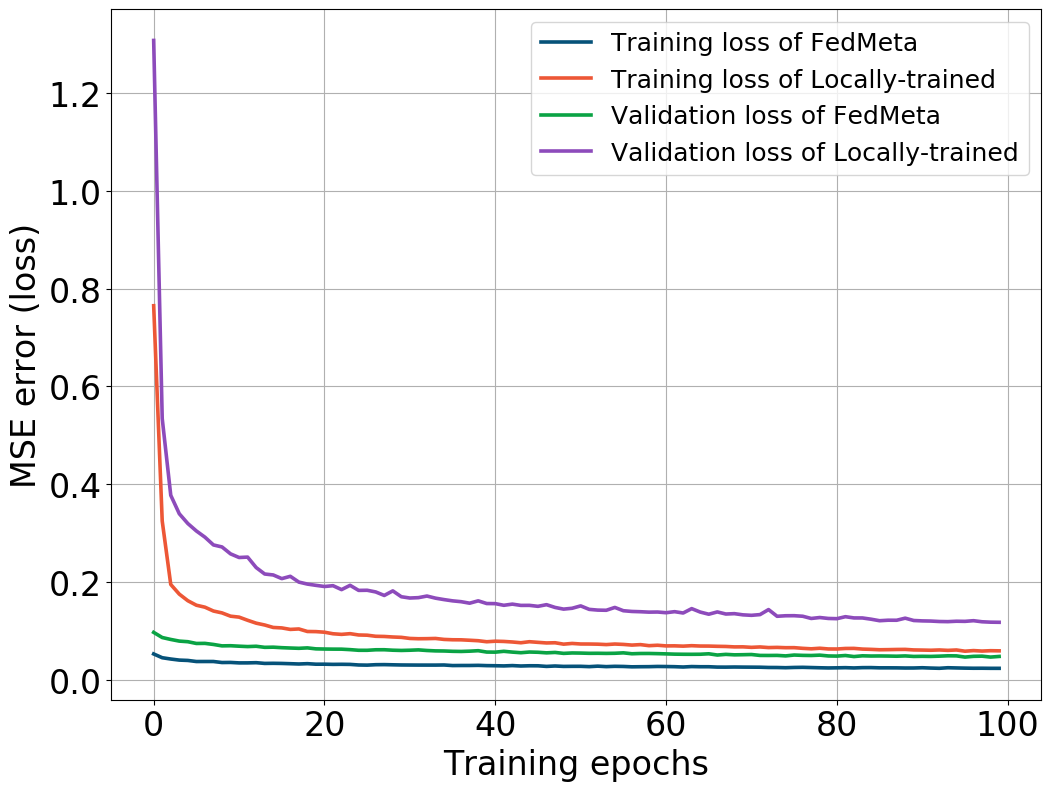}}
\caption{The training and validation losses of locally trained model and model trained with \texttt{FedMeta}. They are trained with the same dataset from a testing task. For \texttt{FedMeta}, here we only present the loss of meta testing step, in which model start training with an initial model trained in former federated learning and meta training steps.}
\label{losscurve}
\end{figure}

For the purpose of validating \texttt{FedMeta}'s fast training and adaptation capability, we compare the MSE error (training/validation loss) of \texttt{FedMeta} with the locally trained model on a randomly chosen task.  As is shown in Fig. \ref{losscurve}, we can clearly observe that with a set of well-trained initial parameters, the model trained with \texttt{FedMeta} has very low training and validation errors even in the first epoch. Moreover, the validation loss of the locally-trained model does not converge after 100 epochs, and it is still greater than that of the model trained with \texttt{FedMeta}. The result shows that \texttt{FedMeta} helps train a generalizable initial model which can be adapted to new tasks quickly.


\vspace{-5pt}
\section{Conclusion and Future Work}
 In this work, we propose a federated learning framework for predicting appliance-level power consumption profiles based on the measurements of aggregated power. The proposed method explicitly integrates meta-learning steps to efficiently adapt to NILM models of different homes and communities, while the resulting learning algorithm eases the communication and computation burden for training a large NILM model. However, our algorithm does not perform well in decomposing the signals of appliances with small amplitudes, as shown in Fig.~\ref{plot3}. Therefore, it is our future research interest to study how to accurately decompose electrical signals with significant amplitude differences. We will also investigate efficient communication and fast converging schemes for learning load disaggregation in a federated and distributed fashion. We will also try to understand theoretically how model adaptation can help each task's energy disaggregation performance.

\vspace{-3pt}
\bibliographystyle{IEEEtran}
\bibliography{bib}

\end{document}